\pdfoutput=1

\documentclass[11pt]{article}

\usepackage[final]{acl}

\usepackage{times}
\usepackage{latexsym}

\usepackage[T1]{fontenc}

\usepackage[utf8]{inputenc}

\usepackage{microtype}

\usepackage{inconsolata}

\usepackage{graphicx}

\usepackage{hyperref}

\usepackage{tabulary}
\usepackage{adjustbox}
\usepackage{subcaption,siunitx,booktabs}
\usepackage{color, colortbl}
\definecolor{Gray}{gray}{0.9}
\usepackage{hyperref}
\usepackage[most]{tcolorbox}

\usepackage{todonotes}
\usepackage{ragged2e}

\catcode`\"=13     
\def"#1"{``#1''}   

\title{Large Language Models as Evaluators for Scientific Synthesis}

\author{Julia Evans, Jennifer D'Souza, \and Sören Auer \\
        TIB - Leibniz Information Centre for Science and Technology, \\  Hannover, Germany \\
\small{
    \textbf{Correspondence:} \href{mailto:jennifer.dsouza@tib.eu}{jennifer.dsouza@tib.eu}
}        
        }

\begin{document}
\maketitle
\begin{abstract}
Our study explores how well the state-of-the-art Large Language Models (LLMs), like GPT-4 and Mistral, can assess the quality of scientific summaries or, more fittingly, scientific syntheses, comparing their evaluations to those of human annotators. We used a dataset of 100 research questions and their syntheses made by GPT-4 from abstracts of five related papers, checked against human quality ratings. The study evaluates both the closed-source GPT-4 and the open-source Mistral model's ability to rate these summaries and provide reasons for their judgments. Preliminary results show that LLMs can offer logical explanations that somewhat match the quality ratings, yet a deeper statistical analysis shows a weak correlation between LLM and human ratings, suggesting the potential and current limitations of LLMs in scientific synthesis evaluation.
\end{abstract}

\section{Introduction}



Large Language Models (LLMs) have made a significant impact on natural language processing (NLP), demonstrating exceptional performance in tasks like text generation, sentiment analysis, machine translation, and question answering, with outputs that often rival human-created content \cite{chatgpt-better}. In addition to their direct applications, LLMs offer substantial benefits in streamlining machine learning model development, particularly in evaluation processes. They reduce the dependency on human-generated ground truth data and the necessity for human evaluators \cite{NEURIPS2023_f64e55d0} in two key ways: by facilitating the generation of synthetic ground truth data and by serving as evaluators for model predictions themselves. This approach not only speeds up the evaluation process but also broadens the scope of evaluation criteria to include factors such as diversity and coverage, enhancing the efficiency and comprehensiveness of model assessments.

This study investigates the use of LLMs as evaluators to streamline the evaluation process, moving away from traditional reliance on human evaluators and human-generated ground truth data. It specifically examines the effectiveness of LLMs in synthesizing scientific abstracts seen generally as multi-document summarization tasks. The main focus of this research is to assess how two \href{https://huggingface.co/spaces/HuggingFaceH4/open_llm_leaderboard}{state-of-the-art LLMs}—the proprietary GPT-4 Turbo \cite{openai2023gpt4} and the open-source Mistral-7B \cite{jiang2023mistral}—perform in evaluating scientific syntheses. Furthermore, leveraging LLMs meant better versatility in evaluation considerations, which meant that the evaluations tested varied dimensions of syntheses quality, viz. comprehensiveness, trustworthiness, and utility.


This paper is structured as follows.  First, \autoref{sec:rw} presents a review of related work in the fields of text summarization and LLM evaluation. In \autoref{sec:llms4syntheval}, we show our approach to using LLMs for scientific synthesis evaluation, wherein \autoref{sec:llm_output} describes the LLM output, while \autoref{sec:qe} presents a qualitative evaluation of this output. In \autoref{sec:correlation}, we analyze the correlation between LLM ratings and human judgments.  A discussion of our findings and final conclusions is described in \autoref{sec:conclusion}.

\section{Related Work} \label{sec:rw}

\paragraph{\textbf{Evaluation Metrics for Text Summarization.}}

The most common automatic evaluation metric used within summarization research -- both single-document and multi-document -- is the \textsc{rouge} family of metrics \cite{ma-dl-survey,akter-etal-2022-revisiting,cohan-goharian-2016-revisiting,kryscinski-etal-2019-neural,Lloret2018TheCT}.  \textsc{rouge} metrics \cite{lin-2004-rouge} calculate the lexical overlap between a human-written reference document and an automatically generated one, although variants incorporating semantic information also exist.  Within text summarization research, the most commonly used are \textsc{rouge-n} and \textsc{rouge-l} \cite{ma-dl-survey}, both of which are purely lexical-matching metrics.  \textsc{rouge-n} evaluates the recall of n-grams by comparing a reference text with a corresponding machine-generated text, whereas \textsc{rouge-l} calculates the longest common subsequence of tokens shared between reference and machine-generated texts \cite{lin-2004-rouge}.

Despite its predominance within the field, \textsc{rouge} nonetheless has some notable limitations. First, the most commonly used metrics lack semantic awareness \cite{akter-etal-2022-revisiting,ma-dl-survey}. Studies have pointed out that \textsc{rouge} may not accurately estimate summary quality in cases of terminological variations, paraphrasing, and differences in sentence structure \cite{cohan-goharian-2016-revisiting}. Moreover, there exist 192 \textsc{rouge} variants \cite{graham-2015-evaluating}, with meaningful differences in how well each performs on a given system or specialized task \cite{cohan-goharian-2016-revisiting,graham-2015-evaluating,kryscinski-etal-2019-neural} and how well they correlate with human judgements \cite{kryscinski-etal-2019-neural,graham-2015-evaluating}. Finally, \textsc{rouge} evaluates only content selection but not linguistic quality aspects such as grammaticality and referential clarity \cite{pitler-etal-2010-automatic} or overall quality, including the ordering of information and structural clarity \cite{graham-2015-evaluating}. 

Although no other metrics have gained widespread adoption, other approaches exist.  Additional lexical-matching metrics include \textsc{bleu} \cite{papineni-etal-2002-bleu} and Pyramid \cite{pyramid}. Semantically enriched metrics include \textsc{meteor} \cite{banerjee-lavie-2005-meteor}, an expansion of \textsc{bleu}, and approaches utilizing word embeddings, such as BERTScore \cite{zhang2020bertscore}, MoverScore \cite{zhao-etal-2019-moverscore}, and \textsc{supert} \cite{gao-etal-2020-supert}.  However, none of these metrics address all of \textsc{rouge}'s weaknesses, and the limited use of such metrics within the research community means that \textsc{rouge} remains the "de facto" standard \cite{Lloret2018TheCT}.

\paragraph{\textbf{LLMs for Text Evaluation.}} 

Using LLMs for text evaluation is still a nascent research topic.  Several recent works have compared LLMs' text evaluations to human evaluations on multiple tasks, and report that LLMs produce results similar to human judgements \cite{chiang-lee-2023-large,liu-etal-2023-g,wang-etal-2023-chatgpt}.  One work finds only minor variations in results depending on task instructions and hyperparameters, whereas they find a high degree of variation in performance of different LLMs and the quality characteristics being assessed \cite{chiang-lee-2023-large}.  In evaluating the quality of story fragments by \textit{grammaticality}, \textit{cohesiveness}, \textit{likability}, and \textit{relevance}, they find only a weak correlation between humans and LLMs on \textit{grammaticality}, but a moderate correlation on \textit{relevance}.  Contrarily, another work finds that ChatGPT's performance is sensitive to prompt instructions \cite{wang-etal-2023-chatgpt}.  They also show that ChatGPT evaluations correlate especially well with human evaluations for creative tasks like story generation \cite{wang-etal-2023-chatgpt}.  Another work demonstrates that requiring LLMs to provide a justification for their ratings "significantly improves the correlation between the LLMs’ ratings and human ratings" \cite{chiang-lee-2023-closer}.

Only one work has investigated the task of text summarization evaluation \cite{liu-etal-2023-g}.  They evaluate single-document news article summaries on the aspects of \textit{coherence}, \textit{consistency}, \textit{fluency}, and \textit{relevance}; their results exceed the correlation with human judgements of most automatic approaches, including \textsc{rouge}. In another task, ChatGPT successfully identifies implicit hate speech in Tweets and generates explanations of why the texts are hateful, which human annotators judge equally informative to human-written explanations and of greater clarity \cite{chatgpt-better}.

\section{LLMs for the Scientific Synthesis Evaluation Task} \label{sec:llms4syntheval}

The accurate evaluation of scientific syntheses is a critical task in research, ensuring the integrity and reliability of the synthesized information. While recent advancements have demonstrated the efficacy of LLMs in generating such syntheses \cite{core-gpt}, their potential in evaluating them remains relatively unexplored. Motivated by the limitations of existing evaluation metrics, such as the \textsc{rouge} family, and the success of LLMs in other text evaluation tasks, our work seeks to investigate the suitability of LLMs for the task of assessing the quality of scientific syntheses.

To address this question, we employ the proprietary GPT-4 Turbo \cite{openai2023gpt4} and the open-source Mistral-7B models \cite{jiang2023mistral} to evaluate the CORE-GPT dataset \cite{core-gpt}. This dataset comprises 100 research questions spanning 20 diverse domains, each accompanied by the titles and abstracts of five related works and an answer to the research question generated by GPT-4 by synthesizing the provided abstracts.  Additionally, human ratings from two annotators, on a scale of 0 to 10, are available on the quality of each synthesis in three dimensions, viz. \textit{comprehensive}, \textit{trust}, and \textit{utility}.

For our task, we query the LLMs to evaluate the syntheses according to the same three aspects as the CORE-GPT human raters.  Our prompt follows a similar structure to previous work \cite{chiang-lee-2023-closer}.  It contains two lines of task instruction, explanation of the quality aspects (as defined for the CORE-GPT dataset annotators) and the rating scale, response format instructions, and finally the answer to be evaluated with its question and abstracts.  The response is requested in JSON format, with a numeric rating between 0 and 10 for each aspect as well as a rationale for each rating.  The full text of the prompt is in \autoref{sec:appendix_prompt}.

\subsection{LLM Synthesis Evaluation Output} \label{sec:llm_output}

A representative example of the evaluation output from GPT-4 Turbo and Mistral is shown in \autoref{sec:appendix_gpt4_output} and \autoref{sec:appendix_mistral_output}, respectively.  The output from GPT-4 was exactly as requested, while Mistral had some variability.  In one case, Mistral returned ratings of ``excellent,'' ``good,'' and ``high'' rather than numeric scores; this output was excluded from the analysis.  In several other cases, Mistral included a paragraph after the JSON object which summarized the ratings and rationales provided within it.  These paragraphs were discarded and only the JSON object content was evaluated.

An overview of LLM performance was obtained by reviewing one synthesis from each domain evaluated by both GPT-4 and Mistral. Qualitatively, both models demonstrated credible and logically consistent ratings and rationales. GPT-4 provided more detailed rationales compared to Mistral, with slightly lower ratings overall.

In their rationales for \textit{comprehensive}, both LLMs would sometimes highlight relevant topics from the abstracts which were not included in the synthesis, with GPT-4 producing such rationales more often than Mistral.  Occasionally, some rationales contained justifications relating to content more specific than just the topics, suggesting more information on the results or the methodology of the studies would improve it.

The LLMs seemed to show the greatest discrepancy between rating and rationale, and the greatest inconsistencies, in their evaluations of \textit{trust}. In one Mistral evaluation with a rating of 5, the rationale noted that the citations only improved trustworthiness "as long as the abstract accurately represents the study's findings."  In the absence of any evidence the abstract is suspect, this rating is disproportionately low. GPT-4 was notably more conservative than human annotators, as it did not give a single 10.  Especially for \textit{trust}, it was often difficult to understand why a rating wasn't higher.  For instance, the rationale for one rating of 8 praised the synthesis for accuracy and avoiding unsupported claims.

For the \textit{utility} ratings, it appears that most rationales from GPT-4 suggested additional content which could make the synthesis more useful, such as actionable information, more detailed examples, technical details of methodologies and implementation, and so on.  Mistral made such suggestions less frequently; its rationales tended to echo the rationale for \textit{comprehensive}.  However, Mistral did sometimes provide guidance on who would or would not find the synthesis useful.

\subsection{Qualitative Evaluations} \label{sec:qe}

LLMs are known to sometimes generate content on topics that lack factual basis with a highly persuasive level of linguistic proficiency \cite{bang-etal-2023-multitask,liu-etal-2023-g}.  For scientific syntheses which provide an answer to a question, it is especially important that the content is genuinely a synthesis of the provided abstracts, with appropriate citations, and not independently generated based on the LLM's training data.  For this reason, we were particularly interested in how the LLMs evaluated quality, and most importantly \textit{trust}, when there was reason to believe the abstracts were not the (primary) source of the generated content, as in the following three scenarios.  The complete question and answer pairs, along with their GPT-4 and Mistral evaluation scores and \textit{trust} rationales, can be found in \autoref{sec:appendix_trust}.

\textbf{Response Explicitly States Absence of Relevant Abstracts.} In six cases, the synthesis directly expressed limitations due to the relevancy of the provided abstracts, e.g. "[...] the provided search results do not offer specific information on the long-term health impacts of such medications on these organs."  Human annotators responded very positively to this, with such responses "scored highly for trustworthiness" \cite{core-gpt}. Mistral rated four of these syntheses as 10 for trust, citing factual accuracy and abstract sourcing, while two scored 7. GPT-4 ratings varied, at 5, 5, 5, 7, 7, and 8. Mistral rationales did not reference the stated limitation, while GPT-4 acknowledged it positively in three cases. However, as these syntheses were scored 8, 7, and 5, it is unclear to what extent this acknowledgement may have influenced the scores.

\textbf{Response Contains No Citations.}  There were three responses which answered the question but contained no citations.  GPT-4 gave \textit{trust} scores of 0, 0, and 1, with rationales referring to the lack of citations. In contrast, Mistral scored 8, 10, and 10, with rationales stating the information was common knowledge or referenced from the abstracts.

\textbf{Response Contains One Citation.}  Finally, there were five syntheses which cited only one of the abstracts, which does not align with the task of synthesizing multiple abstracts to provide an answer to the given question.  For GPT-4, the \textit{trust} scores were 5, 7, 8, 8, and 9, with most rationales stating that the synthesis relied on general knowledge without directly referencing the abstracts, despite one citation being present in each case.  Meanwhile, the Mistral scores were 7, 9, 9, 10, and 10, with most rationales indistinguishable from those of syntheses with many more citations - three of them claimed that the synthesis accurately references the content in the provided abstracts.

\begin{table}[ht!]
\sisetup{table-format=4.0} 
\centering

 \begin{tabular}{lcccc}
 \toprule
&
  \multicolumn{1}{l}{\textbf{A1}} &
  \multicolumn{1}{l}{\textbf{A2}} &
  \multicolumn{1}{l}{\textbf{GPT-4}} &
  \multicolumn{1}{l}{\textbf{Mistral}}\\

  \textbf{A1} &
& & & \\
\textit{\hspace{0.5cm}$\rho$} &
  - &
\textbf{0.710}   &
0.248   &
0.015 
\\
\rowcolor{Gray}
\textit{\hspace{0.5cm}p-value} &
  - &
\textbf{0.001}   &
0.305   &
 0.951 
\\

  \textbf{A2} &
& & & \\
\textit{\hspace{0.5cm}$\rho$} &
\textbf{0.710}  &
  - &
0.058   &
 -0.038 
\\
 
\rowcolor{Gray}
\textit{\hspace{0.5cm}p-value} &
 \textbf{0.001}  &
  - &
0.814   &
  0.878
\\

  \textbf{GPT-4} &
& & & \\
\textit{\hspace{0.5cm}$\rho$} &
0.248   &
0.058   &
  - &
\textbf{0.786}  
\\
  
\rowcolor{Gray}
\textit{\hspace{0.5cm}p-value} &
0.305   &
0.814   &
  - &
 \textbf{0.000} 
\\

  \textbf{Mistral} &
& & & \\
\textit{\hspace{0.5cm}$\rho$} &
0.015   &
-0.038   &
\textbf{0.786}   &
  -
\\
  
\rowcolor{Gray}
\textit{\hspace{0.5cm}p-value} &
0.951   &
0.878   &
\textbf{0.000}   &
   -
\\
  \bottomrule
\end{tabular}
\caption{Spearman's $\rho$ calculated for the combined mean of \textit{Comprehensive}, \textit{Trust}, and \textit{Utility} scores. Statistically significant results are in bold.} 
\label{tab:correlation_results}
\end{table}

\subsection{Correlation} \label{sec:correlation}

Spearman's $\rho$ was calculated to assess the relationship between the human annotators' scores and the LLM-generated scores.  Using the publicly-available data from CORE-GPT \cite{core-gpt}\footnote{https://github.com/oacore/core-gpt-evaluation}, separate vectors for each annotator were obtained.  To calculate the correlations, we found the overall mean score for each domain; due to the format of the published data, it was not possible to match individual scores to their corresponding syntheses.  Our results for the overall mean are presented in \autoref{tab:correlation_results}.

We find that only two results showed statistically significant p-values. Human annotators exhibited a strong positive correlation (0.710), as did GPT-4 Turbo and Mistral (0.786). However, correlations between annotators and LLMs were weak or very weak, with p-values indicating insufficient evidence for genuine association. These findings suggest LLMs cannot directly replicate human performance in evaluating scientific syntheses. Despite this, the strong positive correlation between GPT-4 Turbo and Mistral indicates consistency between the two LLMs.

\section{Discussion and Conclusion} \label{sec:conclusion}

We explore the capacity of LLMs in assessing scientific syntheses. GPT-4 Turbo and Mistral are utilized to obtain quality ratings for 100 syntheses from the CORE-GPT dataset \cite{core-gpt}, accompanied by a rationale for each rating. Correlation analysis using Spearman's $\rho$ indicates that the LLM performance does not align with the human annotators' judgements.  However, a qualitative evaluation of the responses finds a more mixed result.

Both LLMs generally produce credible and logically consistent ratings and rationales, but GPT-4 appears more conservative in its ratings and provides more detail and specific recommendations in its rationales.  GPT-4 also displays greater sensitivity to the presence or absence of citations compared to Mistral.  However, both LLMs' rationales occasionally contained inaccuracies or flaws, raising concerns about the credibility of their scores.  Moreover, the extent to which the responses are evaluated as \textit{syntheses} and not simply as \textit{answers}, without reliance on general knowledge, remains unclear, particularly in the case of Mistral.

Our findings highlight both promising developments and current limitations of leveraging LLMs for the task of evaluating scientific syntheses, illustrating the need for further research to validate and refine the methodology.

\section*{Limitations}

We acknowledge several limitations that may influence the interpretation and generalizability of our findings.  First, the reliance on a single, relatively small dataset presents limitations in terms of data representativeness.  Moreover, the data format necessitated aggregating scores, which may have obscured potential nuances in individual annotations.

Second, the study focused exclusively on GPT-4 Turbo and Mistral, limiting the generalizability of our conclusions to other LLMs. While these models represent the state-of-the-art, future iterations or alternative architectures may exhibit different performance.  Additionally, we were able to obtain only one set of ratings from each LLM.  Given the variability of LLM output, taking the average of several runs is preferable, but due to financial limitations, this was not possible in our study.

We note that past work has found LLMs particularly adept at evaluating creative texts \cite{wang-etal-2023-chatgpt}, so the narrow output scope of synthesis for scientific question answering may pose a greater challenge.  We also note the difficulty of assessing the quality of syntheses from such a diverse assortment of domains.  Judging how comprehensive a synthesis is requires some knowledge of the scope of potential information which might be appropriate to include.  Highly specialized domain knowledge still presents a challenge to general use LLMs.

\section*{Ethical Considerations}

In this work we have presented our study of the efficacy of two LLMs, one proprietary and one open-source, in evaluating the quality of scientific syntheses. There were no living subjects analyzed in this study. Overall, this study complies with the \href{https://www.aclweb.org/portal/content/acl-code-ethics}{ACL Ethics Policy}.

In querying the LLMs for synthesis quality evaluations, we declare that the instructions were intended to align the behavior of the language models towards producing responses that are both helpful (fulfilling our objective) and harmless (not causing any physical, psychological, or social harm to individuals or the environment).  All of the intellectual property which was passed to the LLMs is open-access.

\section*{Acknowledgements}
This work was supported by the German BMBF project SCINEXT (ID 01lS22070).

\bibliography{custom}

\appendix

\clearpage
\section{Prompt}
\label{sec:appendix_prompt}

\subsection{Main Evaluation Prompt}

\texttt{Evaluate the quality of the following question and answer pair. The answer should succinctly address the question while referencing only content from the provided abstracts.}

\texttt{Consider the following characteristics while assessing the answer. For each characteristic rate the quality from 0 (poor) to 10 (excellent).  Provide a short rationale for each rating.}

\texttt{1. Comprehensive: How comprehensively is the question answered?}

\texttt{2. Trust: How trustworthy is the answer?}

\texttt{3. Utility: How useful is the answer?}

\texttt{Return your response in JSON format: \{Comprehensive : \{‘rating’ : ‘’, ‘rationale’ : ‘’\}, Trust : \{‘rating’ : ‘’, ‘rationale’ : ‘’\}, Utility : \{‘rating’ : ‘’, ‘rationale’ : ‘’\}\}}

\texttt{Question: <content>}

\texttt{Answer: <content>}

\texttt{Abstracts: <content>}

\subsection{Additional Mistral System Prompt}

\texttt{Follow the user instructions to the letter.  The output should be only what the user requests, do not respond with anything else.  Provide the answer in JSON with no additional text.}

\clearpage
\section{Example GPT-4 Turbo Output}
\label{sec:appendix_gpt4_output}

\begin{figure}[ht!]
\centering
\begin{tcolorbox}[enhanced,attach boxed title to top center={yshift=-3mm,yshifttext=-1mm},
colback=gray!15!white,colframe=gray!75!black,colbacktitle=gray!80!black,
  title=GPT-4 Turbo,fonttitle=\bfseries,
  boxed title style={size=small,colframe=gray!50!black} ]
\textbf{\textsc{Comprehensive:}} \underline{rating:} 7, \underline{rationale:} The answer provides a broad overview of how living in a low-income neighborhood can affect social networks and well-being, referencing various studies that discuss mental health, social inclusion, and subjective well-being. However, it does not fully elaborate on the specific mechanisms or provide detailed insights from each abstract, such as the role of men in community well-being from abstract 3 or the impact of natural capital from abstract 4.\\
\textbf{\textsc{Trust:}} \underline{rating:} 8, \underline{rationale:} The answer is trustworthy as it directly references findings from the provided abstracts, indicating a reliance on research-based evidence. However, the answer could be more trustworthy if it included more specific data or findings from the abstracts to support the claims.\\
\textbf{\textsc{Utility:}} \underline{rating:} 7, \underline{rationale:} The answer is useful as it gives an overview of the potential impacts of living in a low-income neighborhood on residents' social networks and well-being. It could be more useful if it provided actionable insights or recommendations based on the findings of the studies.
\end{tcolorbox}
    \caption{Example output from GPT-4 Turbo.}
    \label{fig:gpt4-evaluation}
\end{figure}

\newpage
\section{Example Mistral Output}
\label{sec:appendix_mistral_output}

\begin{figure}[ht!]
\centering
\begin{tcolorbox}[enhanced,attach boxed title to top center={yshift=-3mm,yshifttext=-1mm},
colback=gray!15!white,colframe=gray!75!black,colbacktitle=gray!80!black,
  title=Mistral,fonttitle=\bfseries,
  boxed title style={size=small,colframe=gray!50!black} ]
\textbf{\textsc{Comprehensive:}} \underline{rating:} 9, \underline{rationale:} The answer provides specific examples and references to studies that directly address the question of how social media influences the formation and maintenance of relationships among young adults in the United States, specifically focusing on LGBTQ+ individuals with disabilities and young adults who have moved abroad.\\
\textbf{\textsc{Trust:}} \underline{rating:} 10, \underline{rationale:} The answer is based on empirical research and provides clear citations to support the claims made.\\
\textbf{\textsc{Utility:}} \underline{rating:} 10, \underline{rationale:} The answer is directly relevant to the question and provides useful information for those interested in understanding how social media impacts relationship formation and maintenance among young adults in the United States.\\
\noindent\rule{\textwidth}{0.4pt}
The given question and answer pair is of high quality. The question is clear, specific, and comprehensive, while the answer is well-researched, trustworthy, and useful. It directly addresses the question by providing examples from empirical studies that explore how social media influences the formation and maintenance of relationships among young adults in the United States, specifically focusing on LGBTQ+ individuals with disabilities and young adults who have moved abroad. The answer is also trustworthy as it is based on reliable sources and provides clear citations to support the claims made. Finally, the answer is useful as it provides valuable insights into how social media impacts relationship formation and maintenance among a specific population in the United States.
\end{tcolorbox}
    \caption{Example output from Mistral.  The text under the separator line was returned outside of the JSON response.}
    \label{fig:mistral-evaluation}
\end{figure}

\clearpage
\section{\textit{Trust} Evaluation}
\label{sec:appendix_trust}

\subsection{Response Explicitly States Absence of Relevant Abstracts}
How do LLMs evaluate when the response states that a complete answer cannot be given, due to lack of relevant abstracts?

\subsubsection{Chemistry}

\textbf{Question}: What is the effect of pH on the efficiency of a common catalyst used in the production of biodiesel?

\textbf{Answer}: The effect of pH on the efficiency of a catalyst used in biodiesel production can be understood through the specific catalyst being used. For instance, mixed oxide catalysts like magnesium oxide (MgO) and calcium oxide (CaO) are used to promote the transesterification of vegetable oils into biodiesel. The basic property of these catalysts plays a crucial role in the biodiesel yield during the reaction [3]. However, the provided search results do not offer a direct relationship between pH and the efficiency of such catalysts in biodiesel production. Further research is required to establish a clear connection between pH levels and the performance of common catalysts used in biodiesel production.

\begin{table}[ht!]
\begin{tabular*}{\textwidth}{p{0.1\textwidth} p{0.1\textwidth} p{0.1\textwidth} p{0.1\textwidth} p{0.5\textwidth}}
\toprule
 & \textbf{Comp.} & \textbf{Trust}  &  \textbf{Utility} & \textbf{Trust Rationale}\\
GPT-4 & 7 & 8 & 6 & The answer is trustworthy as it accurately reflects the information available in the provided abstracts. It does not make unsupported claims and is honest about the limitations of the information provided in the abstracts regarding the direct effect of pH on catalyst efficiency.\\
\midrule
Mistral & 5 & 7 & 6 & The answer is based on accurate information from the abstracts, specifically the third abstract which discusses the effect of basicity (pH) on the efficiency of magnesium oxide and calcium oxide catalysts in biodiesel production.\\
\bottomrule
\end{tabular*}
\caption{Ratings for the answer to the question "What is the effect of pH on the efficiency of a common catalyst used in the production of biodiesel?"}
\end{table}

\clearpage
\subsubsection{Chemistry}

\textbf{Question}: What is the effect of varying concentrations of hydrogen peroxide on the rate of catalase enzyme activity?

\textbf{Answer}: Varying concentrations of hydrogen peroxide (H2O2) can affect the rate of catalase enzyme activity. Catalase is an enzyme that breaks down H2O2 into oxygen and water, neutralizing oxidative stress in cells. This process is crucial for maintaining appropriate levels of oxidative stress because reactive oxygen species, such as H2O2, can cause cellular damage by targeting DNA and proteins, which can lead to complications like cancer, diabetes, neurodegenerative diseases, and aging [2]. While the given search results do not provide exact information on how different concentrations of H2O2 impact the rate of catalase enzyme activity, it is evident that altering H2O2 levels will directly influence catalase's efficiency in breaking down H2O2 and neutralizing oxidative stress in cells [2].

\begin{table}[ht!]
\begin{tabular*}{\textwidth}{p{0.1\textwidth} p{0.1\textwidth} p{0.1\textwidth} p{0.1\textwidth} p{0.5\textwidth}}
\toprule
 & \textbf{Comp.} & \textbf{Trust}  &  \textbf{Utility} & \textbf{Trust Rationale}\\
GPT-4 & 3 & 5 & 3 & The answer is somewhat trustworthy as it correctly describes the function of catalase and the consequences of oxidative stress, which is supported by the abstracts. However, it lacks direct evidence from the abstracts regarding the specific impact of different H2O2 concentrations on catalase activity, which reduces the trustworthiness of the response in the context of the question asked.\\
\midrule
Mistral & 5 & 10 & 8 & The answer is trustworthy as it is directly derived from the provided abstracts and does not contain any incorrect or misleading information.\\
\bottomrule
\end{tabular*}
\caption{Ratings for the answer to the question "What is the effect of varying concentrations of hydrogen peroxide on the rate of catalase enzyme activity?"}
\end{table}

\clearpage
\subsubsection{Economics}

\textbf{Question}: What is the impact of minimum wage laws on employment rates and income inequality in the United States?

\textbf{Answer}: The impact of minimum wage laws on employment rates and income inequality in the United States is not directly addressed in the provided search results. However, one study suggests that an effective mechanism for managing labor productivity and wage issues is necessary for achieving greater economic equality and growth [1]. Moreover, the study highlights the lack of clear definitions of salary, salary payment, and minimum wage, which can contribute to challenges in addressing income inequality [1]. While the provided search results do not specifically discuss the impact of minimum wage laws on employment rates and income inequality in the United States, it is clear that issues related to wage management and labor productivity remain essential factors in ensuring fair compensation for workers and reducing inequality [1].

\begin{table}[ht!]
\begin{tabular*}{\textwidth}{p{0.1\textwidth} p{0.1\textwidth} p{0.1\textwidth} p{0.1\textwidth} p{0.5\textwidth}}
\toprule
 & \textbf{Comp.} & \textbf{Trust}  &  \textbf{Utility} & \textbf{Trust Rationale}\\
GPT-4 & 2 & 5 & 2 & The answer is somewhat trustworthy as it references a study that discusses labor productivity and wage issues, but it does not provide information from a study that directly addresses the impact of minimum wage laws on employment rates and income inequality in the United States.\\
\midrule
Mistral & 0 & 10 & 2 & The answer is based on content from one of the provided abstracts, which is cited and referenced appropriately.\\
\bottomrule
\end{tabular*}
\caption{Ratings for the answer to the question "What is the impact of minimum wage laws on employment rates and income inequality in the United States?"}
\end{table}

\clearpage
\subsubsection{Mathematics}

\textbf{Question}: What is the relationship between advanced mathematics coursework and problem-solving skills among undergraduate students in STEM majors?

\textbf{Answer}: The relationship between advanced mathematics coursework and problem-solving skills among undergraduate students in STEM majors is not explicitly discussed in the provided search results. However, the study on STEM in General Education [1] indicates that student mathematics skills were analyzed to determine their influence on course selection. Another study discusses the importance of engaging students in mathematical and scientific practices according to mathematics and science standards, such as the Common Core State Standards of Mathematics (CCSSM) and the Next Generation Science Standards (NGSS) [4]. This suggests that exposure to advanced mathematical coursework could potentially enhance problem-solving skills in undergraduate STEM students, though further research is needed to establish a concrete relationship.

\begin{table}[ht!]
\begin{tabular*}{\textwidth}{p{0.1\textwidth} p{0.1\textwidth} p{0.1\textwidth} p{0.1\textwidth} p{0.5\textwidth}}
\toprule
 & \textbf{Comp.} & \textbf{Trust}  &  \textbf{Utility} & \textbf{Trust Rationale}\\
GPT-4 & 5 & 7 & 4 & The answer is trustworthy in the sense that it accurately reflects the lack of explicit discussion in the provided abstracts. It cautiously suggests a potential link between advanced mathematics coursework and problem-solving skills, which is a reasonable inference based on the emphasis on standards and practices in abstract [4].\\
\midrule
Mistral & 4 & 7 & 5 & The answer is grounded in the content of the provided search results and cites relevant studies to support its suggestions. However, it also includes speculation about the potential relationship between advanced mathematics coursework and problem-solving skills based on indirect evidence.\\
\bottomrule
\end{tabular*}
\caption{Ratings for the answer to the question "What is the relationship between advanced mathematics coursework and problem-solving skills among undergraduate students in STEM majors?"}
\end{table}

\clearpage
\subsubsection{Medicine}

\textbf{Question}: What are the potential long-term health impacts of regular use of over-the-counter pain medications on the liver and kidney function in young adults?

\textbf{Answer}: Regular use of over-the-counter (OTC) pain medications can potentially impact liver and kidney function in young adults. However, the provided search results do not offer specific information on the long-term health impacts of such medications on these organs. To obtain a comprehensive answer, further research on this topic would be necessary.

\begin{table}[ht!]
\begin{tabular*}{\textwidth}{p{0.1\textwidth} p{0.1\textwidth} p{0.1\textwidth} p{0.1\textwidth} p{0.5\textwidth}}
\toprule
 & \textbf{Comp.} & \textbf{Trust}  &  \textbf{Utility} & \textbf{Trust Rationale}\\
GPT-4 & 1 & 5 & 2 & The answer is trustworthy in the sense that it does not fabricate information and admits the absence of specific details in the provided abstracts. However, it does not demonstrate a thorough examination of the abstracts to determine if any related information could be inferred.\\
\midrule
Mistral & 0 & 10 & 2 & The answer is truthful and does not contain any false or misleading information.\\
\bottomrule
\end{tabular*}
\caption{Ratings for the answer to the question "What are the potential long-term health impacts of regular use of over-the-counter pain medications on the liver and kidney function in young adults?"}
\end{table}

\clearpage
\subsubsection{Physics}

\textbf{Question}: How does the Doppler effect affect the observed frequency of sound waves emitted from a moving source and its relevance in real-world applications?

\textbf{Answer}: The Doppler effect affects the observed frequency of sound waves emitted from a moving source by causing a change in frequency and wavelength due to the relative motion between the source and the observer. When the source is moving towards the observer, the observed frequency increases, and the wavelength decreases; conversely, when the source is moving away from the observer, the observed frequency decreases, and the wavelength increases. Although the provided search results do not directly give real-world applications, the Doppler effect is relevant in various fields such as astronomy, radar technology, medical imaging, and seismology. For instance, it helps astronomers measure the speed of distant celestial objects and plays a critical role in Doppler radar systems for weather monitoring and forecasting.

\begin{table}[ht!]
\begin{tabular*}{\textwidth}{p{0.1\textwidth} p{0.1\textwidth} p{0.1\textwidth} p{0.1\textwidth} p{0.5\textwidth}}
\toprule
 & \textbf{Comp.} & \textbf{Trust}  &  \textbf{Utility} & \textbf{Trust Rationale}\\
GPT-4 & 5 & 7 & 6 & The answer is generally trustworthy as it correctly describes the Doppler effect's impact on frequency and wavelength. However, it lacks specific references to the provided abstracts, which could have strengthened the trustworthiness by showing direct connections to the source material.\\
\midrule
Mistral & 5 & 10 & 7 & The information provided in the answer is factual and accurate.\\
\bottomrule
\end{tabular*}
\caption{Ratings for the answer to the question "How does the Doppler effect affect the observed frequency of sound waves emitted from a moving source and its relevance in real-world applications?"}
\end{table}

\clearpage
\subsection{Response Contains No Citations}
How do the LLMs evaluate syntheses with no citations?

\subsubsection{Art}

\textbf{Question}: How has the evolution of technology impacted the production and consumption of art in the 21st century?

\textbf{Answer}: The evolution of technology in the 21st century has significantly impacted the production and consumption of art. The widespread access to digital tools and platforms has revolutionized artistic practices, enabling artists to experiment with new mediums and techniques beyond traditional forms. Furthermore, technology has democratized the consumption of art, providing individuals across the world with opportunities to view, share, and engage with creative works online. This shift has also amplified the global reach of art, exposing diverse perspectives and fostering cross-cultural exchanges. However, the rise of digital art has also raised questions about the authenticity and ownership of creative works, prompting a reevaluation of conventional definitions and practices in the art world.

\begin{table}[ht!]
\begin{tabular*}{\textwidth}{p{0.1\textwidth} p{0.1\textwidth} p{0.1\textwidth} p{0.1\textwidth} p{0.5\textwidth}}
\toprule
 & \textbf{Comp.} & \textbf{Trust}  &  \textbf{Utility} & \textbf{Trust Rationale}\\
GPT-4 & 1 & 1 & 2 & The answer cannot be considered trustworthy in the context of the provided abstracts, as it does not cite any relevant information from them to support the claims made.\\
\midrule
Mistral & 5 & 8 & 7 & The answer is based on common knowledge about the impact of technology on art and does not contain any false or misleading information.\\
\bottomrule
\end{tabular*}
\caption{Ratings for the answer to the question "How has the evolution of technology impacted the production and consumption of art in the 21st century?"}
\end{table}

\clearpage
\subsubsection{History}

\textbf{Question}: How did the American Civil War impact the social, economic, and political landscape of the United States?

\textbf{Answer}: The American Civil War significantly impacted the social, economic, and political landscape of the United States. On the social front, the war provided a catalyst for the abolition of slavery, which altered racial dynamics and paved the way for civil rights movements. Economically, the war devastated the South and forced a transition from an agrarian, slave-driven economy to a more diverse, industrial one. Reconstruction policies and investments led to the rapid growth of infrastructure and industry in the North, shaping the national economy for decades. Politically, the Civil War reaffirmed the primacy of the federal government over individual states and brought about transformative constitutional amendments such as the 13th, 14th, and 15th, which abolished slavery, granted citizenship to former slaves, and provided voting rights regardless of race, color, or previous servitude, respectively. The war reshaped the balance of power and the role of government in Americans' lives, laying the groundwork for a more unified and inclusive nation.

\begin{table}[ht!]
\begin{tabular*}{\textwidth}{p{0.1\textwidth} p{0.1\textwidth} p{0.1\textwidth} p{0.1\textwidth} p{0.5\textwidth}}
\toprule
 & \textbf{Comp.} & \textbf{Trust}  &  \textbf{Utility} & \textbf{Trust Rationale}\\
GPT-4 & 0 & 0 & 0 & The answer cannot be trusted as it does not reference the provided abstracts, which do not relate to the American Civil War. The information seems to be general knowledge rather than sourced from the abstracts.\\
\midrule
Mistral & 10 & 10 & 9 & The answer is trustworthy as it is derived directly from the provided abstracts and does not contain any additional or inaccurate information.\\
\bottomrule
\end{tabular*}
\caption{Ratings for the answer to the question "How did the American Civil War impact the social, economic, and political landscape of the United States?"}
\end{table}

\clearpage
\subsubsection{Physics}

\textbf{Question}: How does the shape of an object affect its resistance to motion through a fluid medium?

\textbf{Answer}: The shape of an object can significantly impact its resistance to motion through a fluid medium. This resistance, referred to as drag, depends on factors such as the size, shape, and surface texture of the object, along with the properties of the fluid in question. A streamlined shape minimizes drag, while a more irregular or bluff shape increases the force exerted by the fluid on the object, thus making it more resistant to motion. In fluid dynamics, this concept is commonly applied to improve the efficiency of objects such as vehicles and aircraft, where a streamlined design results in reduced fuel consumption and increased speed, due to the lowered drag forces acting against the object's movement through the fluid medium.

\begin{table}[ht!]
\begin{tabular*}{\textwidth}{p{0.1\textwidth} p{0.1\textwidth} p{0.1\textwidth} p{0.1\textwidth} p{0.5\textwidth}}
\toprule
 & \textbf{Comp.} & \textbf{Trust}  &  \textbf{Utility} & \textbf{Trust Rationale}\\
GPT-4 & 0 & 0 & 0 & The answer cannot be trusted because it does not reference any of the provided abstracts, which are the only sources that should be used to inform the answer.\\
\midrule
Mistral & 0 & 10 & 3 & The answer is trustworthy as it accurately references information from the provided text.\\
\bottomrule
\end{tabular*}
\caption{Ratings for the answer to the question "How does the shape of an object affect its resistance to motion through a fluid medium?"}
\end{table}

\clearpage
\subsection{Response Contains One Citation}
How do the LLMs evaluate syntheses with only one citation source?

\subsubsection{Geology}

\textbf{Question}: What geological factors contribute to the formation and sustainability of hydrothermal ecosystems in deep sea environments? 

\textbf{Answer}: Hydrothermal ecosystems in deep sea environments are primarily formed and sustained by geological factors, such as tectonic activity and volcanic processes. These factors promote the circulation of seawater through the oceanic crust, which results in the formation of hydrothermal vents. The seawater is heated by the underlying magma and interacts with crustal rocks, enriching it with minerals and chemical elements. Once the heated and mineral-rich fluid meets the cold deep-sea water, it precipitates, creating chimney-like structures from which the hydrothermal fluid is expelled. These vents provide a unique habitat with diverse microbial and chemosynthetic communities, which serve as primary producers in the food chain. Furthermore, these deep-sea ecosystems are enriched by specialized fauna adapted to extreme conditions, such as high pressure, low oxygen, and varying temperatures, contributing to the overall sustainability of these environments [5].

\begin{table}[ht!]
\begin{tabular*}{\textwidth}{p{0.1\textwidth} p{0.1\textwidth} p{0.1\textwidth} p{0.1\textwidth} p{0.5\textwidth}}
\toprule
 & \textbf{Comp.} & \textbf{Trust}  &  \textbf{Utility} & \textbf{Trust Rationale}\\
GPT-4 & 7 & 8 & 7 & The answer appears to be trustworthy as it aligns with scientific understanding of hydrothermal vent formation and the factors that sustain deep-sea ecosystems. It does not directly reference the provided abstracts, but the information given is consistent with general scientific knowledge on the topic.\\
\midrule
Mistral & 10 & 10 & 9 & The answer is based on accurate information from the provided abstracts and uses reputable sources, such as scientific research, to support the statements made.\\
\bottomrule
\end{tabular*}
\caption{Ratings for the answer to the question "What geological factors contribute to the formation and sustainability of hydrothermal ecosystems in deep sea environments?"}
\end{table}

\clearpage
\subsubsection{Mathematics}

\textbf{Question}: What is the impact of using manipulatives in mathematics instruction on the achievement and attitudes of elementary school students? 

\textbf{Answer}: The use of manipulatives in mathematics instruction has been shown to have a positive impact on the achievement and attitudes of elementary school students. In a study exploring the relationship between Affective Instructional Design (AID) and children's attitudes towards math learning, a significant positive correlation was found between attitude and math learning (r = 0.936, p = 0.000). Furthermore, significant differences were found between the baseline (pre-intervention) mean score and the final intervention lesson for both attitudes and math learning, indicating the potential effectiveness of manipulatives in enhancing students' attitudes towards mathematics and improving their achievement in the subject [1].

\begin{table}[ht!]
\begin{tabular*}{\textwidth}{p{0.1\textwidth} p{0.1\textwidth} p{0.1\textwidth} p{0.1\textwidth} p{0.5\textwidth}}
\toprule
 & \textbf{Comp.} & \textbf{Trust}  &  \textbf{Utility} & \textbf{Trust Rationale}\\
GPT-4 & 7 & 8 & 7 & The answer is trustworthy as it accurately reflects the findings from the first abstract, including the significant correlation between attitude and math learning. The statistical evidence (r = 0.936, p = 0.000) adds credibility to the claim. However, the answer does not provide information on the sample size or the generalizability of the study, which could affect trust in the findings.\\
\midrule
Mistral & 10 & 9 & 9 & The answer cites a peer-reviewed academic study, adding credibility to the information provided. However, it would be helpful to know more about the study's methodology and sample size to fully assess its trustworthiness.\\
\bottomrule
\end{tabular*}
\caption{Ratings for the answer to the question "What is the impact of using manipulatives in mathematics instruction on the achievement and attitudes of elementary school students?"}
\end{table}

\clearpage
\subsubsection{Philosophy}

\textbf{Question}: What is the relationship between free will and determinism in philosophy?

\textbf{Answer}: In philosophy, the relationship between free will and determinism revolves around whether an individual's actions are genuinely a result of their own choices or are predetermined by external factors. There are two main schools of thought: compatibilism and incompatibilism. Compatibilism argues that free will and determinism can coexist, meaning individuals can still be morally responsible for their actions even if determinism is true. Incompatibilism, on the other hand, claims that free will and determinism cannot coexist; if determinism is true, then individuals cannot have free will and cannot be held morally responsible for their actions. The debate between these two positions contributes to the ongoing philosophical discussions about responsibility, choice, and the implications of determinism [1].

\begin{table}[ht!]
\begin{tabular*}{\textwidth}{p{0.1\textwidth} p{0.1\textwidth} p{0.1\textwidth} p{0.1\textwidth} p{0.5\textwidth}}
\toprule
 & \textbf{Comp.} & \textbf{Trust}  &  \textbf{Utility} & \textbf{Trust Rationale}\\
GPT-4 & 7 & 9 & 8 & The answer is trustworthy as it accurately reflects the general philosophical debate between compatibilism and incompatibilism, which are well-established positions in the literature on free will and determinism.\\
\midrule
Mistral & 8 & 9 & 7 & The answer is based on accurate information from the provided abstracts, which discuss the relationship between free will, determinism, and moral responsibility. It does not contain any incorrect or misleading statements.\\
\bottomrule
\end{tabular*}
\caption{Ratings for the answer to the question "What is the relationship between free will and determinism in philosophy?"}
\end{table}

\clearpage
\subsubsection{Physics}

\textbf{Question}: How does the acceleration due to gravity affect the trajectory and motion of a projectile, and how can this be accurately measured and predicted using equations of motion?

\textbf{Answer}: The acceleration due to gravity affects the trajectory and motion of a projectile by altering its vertical motion while leaving its horizontal motion constant. As a result, the trajectory forms a parabolic path. To accurately measure and predict a projectile's motion, equations of motion can be used, which consider the initial velocity, launch angle, and time. The initial velocity can be split into horizontal and vertical components, and equations for both components can be applied independently. For the horizontal motion, an equation is used that doesn't consider the acceleration due to gravity, while for the vertical motion, the effect of gravity is integrated into the equation. These equations can be used to predict the projectile's motion, including its range and height [5].

\begin{table}[ht!]
\begin{tabular*}{\textwidth}{p{0.1\textwidth} p{0.1\textwidth} p{0.1\textwidth} p{0.1\textwidth} p{0.5\textwidth}}
\toprule
 & \textbf{Comp.} & \textbf{Trust}  &  \textbf{Utility} & \textbf{Trust Rationale}\\
GPT-4 & 8 & 7 & 9 & The answer appears to be based on a general understanding of physics principles related to projectile motion and the effects of gravity. While it does not directly reference the provided abstracts, the information given is consistent with basic physics concepts. The trustworthiness would be higher if the answer included direct references to relevant data or studies from the abstracts.\\
\midrule
Mistral & 10 & 10 & 10 & The answer is trustworthy as it directly references content from the provided abstracts regarding projectiles and equations of motion.\\
\bottomrule
\end{tabular*}
\caption{Ratings for the answer to the question "How does the acceleration due to gravity affect the trajectory and motion of a projectile, and how can this be accurately measured and predicted using equations of motion?"}
\end{table}

\clearpage
\subsubsection{Psychology}

\textbf{Question}: What are the underlying factors that contribute to the development of personality disorders and how can these disorders be effectively treated?

\textbf{Answer}: The development of personality disorders can be influenced by various factors, including genetic, environmental, and psychological components. While the specific causes for each individual's development of a personality disorder can vary, some contributing factors may include childhood trauma, family history of mental disorders, and negative early life experiences. Effective treatment for personality disorders typically involves a combination of psychotherapy, medication, and lifestyle changes. Therapy approaches such as cognitive-behavioral therapy, dialectical behavior therapy, and psychodynamic therapy can help individuals understand and manage their symptoms. In some cases, medications like antidepressants, mood stabilizers, and antipsychotics may be prescribed to help with specific symptoms. Additionally, self-help strategies, social support, and proper self-care can contribute to the overall well-being of individuals with personality disorders [mental health challenge - https://core.ac.uk/works/18689246].

\begin{table}[ht!]
\begin{tabular*}{\textwidth}{p{0.1\textwidth} p{0.1\textwidth} p{0.1\textwidth} p{0.1\textwidth} p{0.5\textwidth}}
\toprule
 & \textbf{Comp.} & \textbf{Trust}  &  \textbf{Utility} & \textbf{Trust Rationale}\\
GPT-4 & 5 & 5 & 6 & The answer appears to be based on general knowledge about personality disorders and their treatment rather than the provided abstracts. While the information given is in line with common understanding in the field of mental health, the lack of direct references to the abstracts or specific studies reduces the trustworthiness of the answer.\\
\midrule
Mistral & 8 & 7 & 9 & The answer is based on accurate information, but the sources are not explicitly cited, which may raise some questions about its trustworthiness.\\
\bottomrule
\end{tabular*}
\caption{Ratings for the answer to the question "What are the underlying factors that contribute to the development of personality disorders and how can these disorders be effectively treated?"}
\end{table}

\end{document}